\documentclass[11pt]{article}
\pdfoutput=1

\usepackage{acl}
\usepackage{times}
\usepackage{latexsym}
\usepackage[T1]{fontenc}
\usepackage[utf8]{inputenc}
\usepackage{microtype}
\usepackage{inconsolata}
\usepackage{booktabs}
\usepackage{amsmath}
\usepackage{graphicx}
\usepackage{multirow}
\usepackage{xcolor}

\title{Caraman at SemEval-2026 Task 8: Three-Stage Multi-Turn Retrieval\\with Query Rewriting, Hybrid Search, and Cross-Encoder Reranking}

\author{David-Maximilian Caraman \\
  Babe\c{s}-Bolyai University \\
  Cluj-Napoca, Romania \\
  \texttt{david.caraman@stud.ubbcluj.ro} \And
  Gheorghe Cosmin Silaghi \\
  Babe\c{s}-Bolyai University \\
  Cluj-Napoca, Romania \\
  \texttt{gheorghe.silaghi@ubbcluj.ro}}

\begin{document}
\maketitle

% =============================================================================
% ABSTRACT
% =============================================================================
\begin{abstract}
We describe our system for SemEval-2026 Task~8 (MTRAGEval), participating in Task~A (Retrieval) across four English-language domains.
Our approach employs a three-stage pipeline: (1)~query rewriting via a LoRA-fine-tuned Qwen~2.5 7B model that transforms context-dependent follow-up questions into standalone queries, (2)~hybrid BM25 and dense retrieval combined through Reciprocal Rank Fusion, and (3)~cross-encoder reranking with BGE-reranker-v2-m3.
On the official test set, the system achieves nDCG@5 of 0.531, ranking 8th out of 38 participating systems and 10.7\% above the organizer baseline.
Development comparisons reveal that domain-specific temperature tuning for query generation, where technical domains benefit from deterministic decoding and general domains from controlled randomness, provides consistent gains, while more complex strategies such as domain-aware prompting and multi-query expansion degrade performance.
\end{abstract}

% =============================================================================
% 1. INTRODUCTION
% =============================================================================
\section{Introduction}

Multi-turn conversational retrieval poses a fundamental challenge for information retrieval systems: users formulate follow-up questions that depend on prior conversational context through pronouns (\textit{``How much does it cost?''}), ellipsis, and implicit topic continuations.
A retrieval system that processes only the latest user utterance loses critical context leads to severe performance degradation in later conversation turns \citep{10.1162/TACL.a.19}.

SemEval-2026 Task~8 \citep{Rosenthal2026MTRAGEval} provides a rigorous benchmark for this problem, spanning four English domains with 777 retrieval queries drawn from 110 multi-turn human conversations. Task A requires participants to produce passages that are relevant to the user’s final question, while evaluation being conducted only on the subset of answerable questions.

We address this challenge with a three-stage pipeline where each component targets a distinct failure mode: 
(i) \textbf{query rewriting} resolves conversational dependencies, (ii) \textbf{hybrid retrieval} captures both lexical and semantic matches, and (iii) \textbf{cross-encoder reranking} refines the final ranking through fine-grained query-passage interaction.
The query rewriter is a Qwen~2.5 7B Instruct model \citep{yang2024qwen25} fine-tuned with LoRA \citep{hu2022lora} on the MTRAGEval gold rewrites, trained entirely on Apple~Silicon using the MLX framework.\footnote{\url{https://github.com/ml-explore/mlx}}

Development comparison studies presented in this paper surface instructive negative results: domain-aware prompting and multi-query expansion both \emph{degraded} performance relative to simple rewriting; fine-tuning the cross-encoder reranker on task-specific data yielded only marginal gains; and larger candidate pools for reranking \emph{decreased} quality, a counterintuitive finding we analyze in detail.
Our code\footnote{\url{https://github.com/davidcaraman/semeval2026-mtrag-retrieval}} and fine-tuned model\footnote{\url{https://huggingface.co/caraman/Qwen2.5-7B-mtrag-query-rewriter-final}} are publicly available.

% =============================================================================
% 2. BACKGROUND
% =============================================================================
\section{Background}

\subsection{Task Description}

MTRAGEval \citep{Rosenthal2026MTRAGEval}, built on the MTRAG-UN benchmark \citep{rosenthal2026mtragunbenchmarkopenchallenges}, evaluates retrieval-augmented generation in multi-turn conversational settings.
Task~A (Retrieval) requires systems to return a ranked list of 10 passages per query from domain-specific corpora totaling 366,479 passages across 78,170 documents in four English domains: \textbf{ClapNQ} (Wikipedia, 183K passages), \textbf{Cloud} (IBM technical documentation, 72K), \textbf{FiQA} (financial forum discussions, 61K), and \textbf{Govt} (U.S.\ government policy documents, 50K).
Each corpus uses 512-token passages with 100-token overlap.
The primary evaluation metric is nDCG@5 \citep{jarvelin2002ndcg}, with nDCG@10 and Recall@10 as secondary measures.

Relevance judgments are derived from human annotations. Following the MTRAG-UN taxonomy \citep{rosenthal2026mtragunbenchmarkopenchallenges}, the benchmark contains a substantial fraction of unanswerable and underspecified queries; these remain undisclosed in the test set and are excluded from scoring.

\subsection{Related Work}

Our system builds on three lines of work.
\textbf{Conversational query rewriting} transforms context-dependent questions into standalone queries suitable for traditional retrievers \citep{vakulenko2021question}; we extend this with domain-specific temperature control for the rewriter's generation.
\textbf{Hybrid retrieval} combines lexical matching (BM25; \citealp{robertson2009bm25}) with dense semantic search \citep{reimers2019sbert} through fusion methods such as Reciprocal Rank Fusion \citep{cormack2009rrf}.
\textbf{Cross-encoder reranking} \citep{nogueira2020passage} jointly encodes query-passage pairs for fine-grained relevance scoring, consistently outperforming bi-encoder approaches at the cost of higher latency.

% =============================================================================
% 3. SYSTEM OVERVIEW
% =============================================================================
\section{System Overview}

Our pipeline processes each conversational query through three sequential stages.
We describe each stage's design, the experimental evidence motivating it, and the key hyperparameters needed for replication.

\subsection{Stage 1: Query Rewriting}
\label{sec:rewriting}

Multi-turn queries frequently contain unresolved references that make them unintelligible to a retrieval system in isolation.
For instance, given a conversation about IBM Cloud where the user asks \textit{``How much does it cost?''}, the pronoun \textit{``it''} must be resolved to produce the standalone query \textit{``What is the pricing for IBM Cloud services?''}.

This stage takes the base Qwen~2.5 7B Instruct model, fine-tunes it with LoRA on gold rewrites from the MTRAGEval training set, and applies a fixed prompt template (Appendix~\ref{app:prompt}) that provides conversation history and instructs the model to produce a standalone query.
At inference time, each query is rewritten with a domain-specific temperature identified through a systematic sweep on the holdout set (Section~\ref{sec:rewriting_analysis}), producing one rewritten query per domain corpus.

\paragraph{Model and Training.}
We fine-tune Qwen~2.5 7B Instruct \citep{yang2024qwen25} using LoRA \citep{hu2022lora} with the parameters presented in Appendix~\ref{app:hyperparams}. 
% rank~16, $\alpha{=}32$, and dropout 0.15, applied to all attention projections (\texttt{q/k/v/o\_proj}) and MLP layers (\texttt{gate/up/down\_proj}) across all 28 transformer layers, yielding 40.4M trainable parameters (0.53\% of the total 7.6B).
These hyperparameters were selected through systematic experimentation over LoRA dropout (0.05--0.15), learning rate ($5{\times}10^{-6}$--$1{\times}10^{-5}$), and the number of adapted layers (16--28), with the final configuration chosen based on holdout validation loss.
Training uses the MLX framework on an Apple M4~Max (128\,GB unified memory) with AdamW ($\text{lr}{=}10^{-5}$, weight decay 0.01), effective batch size 16 (micro-batch~2, gradient accumulation~8), and gradient checkpointing for memory efficiency.
The model trains for 500 iterations on 699 query rewriting examples derived from MTRAGEval gold rewrites, with 78 examples held out for validation.

\begin{figure}[!t]
\centering
\includegraphics[width=\columnwidth]{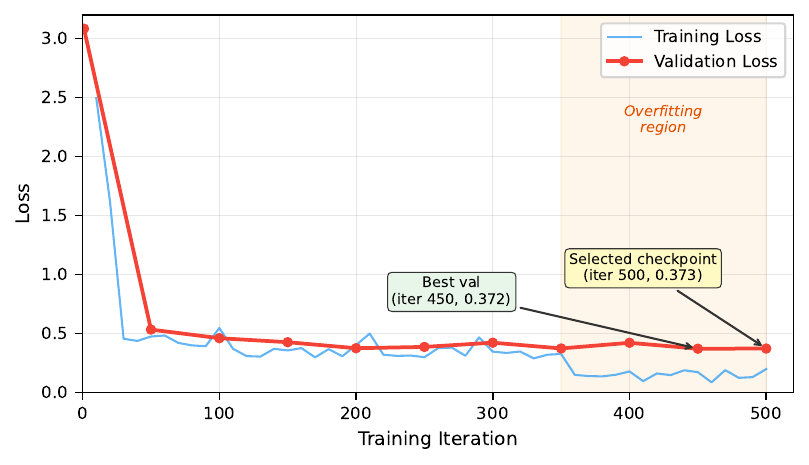}
\caption{LoRA fine-tuning loss curve for the query rewriter.
% Validation loss plateaus around 0.37 after iteration~200, while training loss continues to decrease after iteration~350, indicating mild overfitting. The selected checkpoint (iteration~500) has validation loss within 0.001 of the best.
}
\label{fig:training_curve}
\end{figure}

\paragraph{Checkpoint Selection.}
Figure~\ref{fig:training_curve} shows the training dynamics.
Validation loss decreases rapidly in the first 200 iterations (3.084 $\to$ 0.376), then plateaus around 0.37--0.42.
After iteration~350, training loss drops sharply to ${\sim}$0.1 while validation loss oscillates, indicating the onset of overfitting.
We select the final checkpoint at iteration~500 (validation loss 0.373), which is within 0.001 of the best validation loss (0.372 at iteration~450), confirming that the model has converged without significant overfitting despite the low training loss.

The prompt template provides up to 10 turns of conversation history followed by the current question, instructing the model to resolve pronouns, include necessary context, and produce a concise, search-friendly standalone query (see Appendix~\ref{app:prompt} for the full prompt).

\paragraph{Domain-Specific Temperature.}
Because optimal rewriting behavior varies across domains, our system generates separate rewritten queries for each domain using domain-specific temperatures, producing four queries per conversation turn, one per domain corpus.
We conducted a systematic temperature sweep (0.0--1.0) on the development holdout to identify the optimal temperature per domain; the full results and analysis are presented in Section~\ref{sec:rewriting_analysis}.

\paragraph{First-Turn Optimization.}
First-turn queries have no conversational history and are therefore already standalone.
We validate empirically on 19 first-turn holdout queries that skipping the rewriter produces identical nDCG@5 (0.381) while reducing inference time by 29\% (32\,s vs.\ 45\,s per domain).

\subsection{Stage 2: Hybrid Retrieval}
\label{sec:retrieval}

No single retrieval paradigm dominates across all domains: lexical matching captures exact technical terms critical for Cloud, while semantic similarity handles paraphrases prevalent in ClapNQ.
We combine both approaches through score-level fusion.

\paragraph{BM25 (Lexical).}
We use BM25S \citep{lu2024bm25s, robertson2009bm25} with English stopword removal and no stemming.
Omitting stemming preserves exact technical terms (e.g., \textit{``Kubernetes''}, \textit{``401(k)''}) that stemming would corrupt.
Each query retrieves 50 candidates.

\paragraph{Dense (Semantic).}
Queries and passages are encoded with BGE-base-en-v1.5 \citep{chen-etal-2024-m3} (768 dimensions) and searched against FAISS IndexFlatIP indices \citep{johnson2019faiss} with L2-normalized embeddings (cosine similarity).
Each query retrieves 50 candidates.

\paragraph{Reciprocal Rank Fusion.}
BM25 and dense result lists are merged using RRF \citep{cormack2009rrf}:
\begin{equation}
\text{score}_{\text{RRF}}(d) = \sum_{r \in \{B, D\}} \frac{1}{k + \text{rank}_r(d)}
\label{eq:rrf}
\end{equation}
where $k{=}60$, $B$ is the BM25 ranking, and $D$ is the dense ranking.
RRF requires no learned weights and produces a single fused candidate list.
Documents appearing in both lists receive contributions from both rankings, naturally boosting passages with cross-modal agreement.

\subsection{Stage 3: Cross-Encoder Reranking}
\label{sec:reranking}

The fused candidate list is reranked by BGE-reranker-v2-m3 \citep{chen-etal-2024-m3}, a 568M-parameter cross-encoder that jointly encodes each query-passage pair.
Table~\ref{tab:rerankers} compares four reranker models at $k{=}50$ candidates.
BGE-reranker-v2-m3 (nDCG@5 0.416) substantially outperforms both ms-marco-MiniLM-L-12-v2 (0.375) and BGE-reranker-base (0.307).
We also evaluated IBM Granite Reranker R2 \citep{granite2025embedding}, which achieved nDCG@5 of 0.391; we selected BGE-v2-m3 for the final system as it achieved the highest score across all candidate pool sizes (Table~\ref{tab:candidates}).

\begin{table}[!t]
\centering
\small
\begin{tabular}{lcc}
\toprule
\textbf{Reranker} & \textbf{Params} & \textbf{nDCG@5} \\
\midrule
BGE-reranker-base & 110M & 0.307 \\
ms-marco-MiniLM-L-12-v2 & 33M & 0.375 \\
IBM Granite R2 & 149M & 0.391 \\
\textbf{BGE-reranker-v2-m3} & \textbf{568M} & \textbf{0.416} \\
\bottomrule
\end{tabular}
\caption{Cross-encoder reranker comparison at $k{=}50$ candidates (nDCG@5, no query rewriting).}
\label{tab:rerankers}
\end{table}

\paragraph{Candidate Pool Size.}
Reranking candidates ($k$) involve a trade-off between recall and noise from borderline passages.
Our sweep over $k \in \{30, 50, 100, 250, 500\}$ with four rerankers (Table~\ref{tab:candidates}) shows that all models degrade from their peak to $k{=}500$, but the optimal $k$ is model-dependent: three of four rerankers peak at $k{=}30$, yet BGE-v2-m3 achieves its best nDCG@5 at $k{=}50$ (0.416 vs.\ 0.411 at $k{=}30$), indicating that stronger cross-encoders can exploit a moderately larger pool before noise dominates.
We adopt $k{=}50$ for the final system.

\begin{table}[!t]
\centering
\small
\setlength{\tabcolsep}{3pt}
\resizebox{\columnwidth}{!}{%
\begin{tabular}{lccccc}
\toprule
\textbf{Reranker} & $k{=}30$ & $k{=}50$ & $k{=}100$ & $k{=}250$ & $k{=}500$ \\
\midrule
BGE-v2-m3       & .411 & \textbf{.416} & .400 & .390 & .388 \\
IBM Granite R2   & .399 & .391          & .392 & .385 & .380 \\
ms-marco-MiniLM  & .376 & .375          & .373 & .369 & .366 \\
BGE-base         & .321 & .307          & .290 & .266 & .251 \\
\bottomrule
\end{tabular}%
}
\caption{Effect of candidate pool size on reranking quality (nDCG@5). Three of four models peak at $k{=}30$, but the selected model (BGE-v2-m3) peaks at $k{=}50$.}
\label{tab:candidates}
\end{table}

We also evaluated fine-tuning BGE-reranker-v2-m3 on task-specific hard negatives (passages from ranks 20--100).
The gain was marginal: nDCG@5 improved from 0.396 to 0.402 (+1.5\%), while Recall@10 actually declined from 0.544 to 0.526.
The pre-trained model is already well-calibrated for this mixed-domain distribution.

% =============================================================================
% 4. EXPERIMENTAL SETUP
% =============================================================================
\section{Experimental Setup}

\paragraph{Data.}
We split the 110 human conversations from the MTRAGEval training set into 88 training conversations (613 queries) and 22 holdout conversations (164 queries: ClapNQ~27, Cloud~47, Govt~62, FiQA~28).
All development results reported in this paper are on the holdout set.
Query rewriting training data comprises 699 training and 78 validation examples extracted from gold rewrites.
For the final submission, the query rewriter was retrained on all 777 examples (training and holdout combined) using the same hyperparameters selected during development, since the holdout set was no longer needed for model selection.
Official competition results are on the separate test set released by the organizers.

\paragraph{Metrics.}
We report nDCG@5 \citep{jarvelin2002ndcg} as the primary metric (matching the official evaluation), with nDCG@10 and Recall@10 as secondary measures, computed using \texttt{pytrec\_eval} \citep{VanGysel2018pytreceval}.

\paragraph{Infrastructure.}
All experiments run on an Apple M4~Max with 128\,GB unified memory.
LoRA training uses the MLX framework;\footnote{\texttt{mlx} v0.12+, \texttt{mlx-lm} v0.12+} retrieval and reranking use PyTorch with MPS acceleration.\footnote{Key libraries: \texttt{bm25s} v0.2+, \texttt{sentence-transformers} v2.2+, \texttt{faiss-cpu} v1.7.4+, \texttt{FlagEmbedding} v1.2+, \texttt{transformers} v4.36+.}

% =============================================================================
% 5. RESULTS AND ANALYSIS
% =============================================================================
\section{Results and Analysis}

\subsection{Query Rewriting Analysis}
\label{sec:rewriting_analysis}

To identify the optimal generation temperature for each domain, we performed a systematic end-to-end sweep over seven temperature values (0.0, 0.1, 0.2, 0.3, 0.5, 0.7, 1.0) on the 164-query development holdout set.
For each temperature, we ran the full pipeline (rewriting $\to$ hybrid retrieval $\to$ reranking) and measured nDCG@5, ensuring that the selected temperatures optimize the final retrieval quality rather than an intermediate proxy.
Table~\ref{tab:tempsweep} presents the results.

Table~\ref{tab:perdomain_full} provides the full per-domain breakdown at each domain's optimal temperature on the development holdout.

\begin{table}[!t]
\centering
\small
\begin{tabular}{lccccc}
\toprule
\textbf{Temp} & \textbf{Overall} & \textbf{ClapNQ} & \textbf{Cloud} & \textbf{FiQA} & \textbf{Govt} \\
\midrule
None & .371 & .535 & .432 & .275 & .296 \\
\midrule
0.0 & .416 & .550 & \textbf{.473} & .321 & .358 \\
0.1 & .408 & .542 & .472 & .290 & .354 \\
\textbf{0.2} & \textbf{.422} & \textbf{.563} & .468 & .321 & .371 \\
0.3 & .416 & .544 & .441 & \textbf{.346} & \textbf{.373} \\
0.5 & .403 & .524 & .463 & .277 & .362 \\
0.7 & .390 & .527 & .438 & .280 & .343 \\
1.0 & .380 & .522 & .440 & .233 & .338 \\
\bottomrule
\end{tabular}
\caption{Temperature sweep for query rewriting (nDCG@5) on the development holdout. ``None'' denotes the no-rewriting baseline using last-turn queries. Bold indicates best per column.}
\label{tab:tempsweep}
\end{table}

\begin{table}[!t]
\centering
\scriptsize
\begin{tabular}{p{1.1cm}cccc}
\toprule
\textbf{Domain} & \textbf{Temp} & \textbf{nDCG@5} & \textbf{nDCG@10} & \textbf{Recall@10} \\
\midrule
ClapNQ  & 0.2 & 0.563 & 0.593 & 0.710 \\
Cloud   & 0.0 & 0.473 & 0.518 & 0.599 \\
Govt    & 0.3 & 0.373 & 0.438 & 0.551 \\
FiQA    & 0.3 & 0.346 & 0.367 & 0.439 \\
\midrule
Overall ($t{=}0.2$) & 0.2 & 0.422 & 0.467 & 0.569 \\
\bottomrule
\end{tabular}
\caption{Per-domain results at domain-optimal temperatures on the development holdout (164 queries).}
\label{tab:perdomain_full}
\end{table}

Query rewriting at the best uniform temperature ($t{=}0.2$) improves nDCG@5 from 0.371 (no rewriting) to 0.422, a \textbf{13.7\%} relative gain, confirming that resolving conversational dependencies is the single most impactful intervention for multi-turn retrieval.

The optimal temperature varies substantially across domains.
\textbf{Cloud} (technical documentation) achieves its best result at $t{=}0.0$ (0.473): its precise technical vocabulary (e.g., \textit{``Kubernetes''}, \textit{``VPC peering''}) means any generation randomness risks substituting incorrect technical terms, corrupting the lexical signal that BM25 relies on.
\textbf{ClapNQ} (Wikipedia) peaks at $t{=}0.2$ (0.563), reflecting its well-structured language that benefits from minimal reformulation diversity.
\textbf{FiQA} (financial forums) and \textbf{Govt} (government documents) both prefer $t{=}0.3$ (0.346 and 0.373), as their more ambiguous query patterns, including informal forum language in FiQA and policy-specific terminology in Govt, benefit from slightly more exploratory rewriting.

Performance degrades monotonically above $t{=}0.3$ across all domains, with $t{=}1.0$ reducing overall nDCG@5 to 0.380, barely above the no-rewriting baseline of 0.371.
FiQA is particularly sensitive: at $t{=}1.0$, its nDCG@5 drops to 0.233, \emph{below} the no-rewriting baseline (0.275), as high randomness corrupts its already-ambiguous financial jargon.
This observed correlation between domain formality and optimal temperature, where technical domains favor deterministic generation while informal domains benefit from controlled randomness, confirms that query rewriting demands high precision and motivates the per-domain temperature configuration adopted in our final system.

\subsection{Comparison Studies}

\paragraph{Query Strategy Comparison.}
We tested two alternatives to simple rewriting.
\textbf{Domain-aware prompting} injects domain metadata into the rewriter prompt (e.g., \textit{``This is a technical cloud computing query''}). This degraded nDCG@5 to 0.350, \emph{below} even the no-rewriting baseline of 0.371.
The domain context caused the model to over-specialize queries, narrowing retrieval scope and introducing domain-specific jargon not present in the original question.
\textbf{Multi-query expansion} generates three query variants and merges their retrieved results. This also scored 0.350, as the additional queries diluted the signal from the primary rewrite with noisy alternatives.

Table~\ref{tab:strategies}  provides the full comparison.
Both results demonstrate that added complexity in query formulation does not compensate for a well-tuned simple rewriter.

\begin{table}[!t]
\centering
\small
\begin{tabular}{lcc}
\toprule
\textbf{Strategy} & \textbf{nDCG@5} & \textbf{nDCG@10} \\
\midrule
No rewriting (baseline) & 0.371 & 0.412 \\
\textbf{Simple rewriting ($t{=}0.2$)} & \textbf{0.422} & \textbf{0.467} \\
Domain-aware prompting & 0.350 & 0.399 \\
Multi-query expansion & 0.350 & 0.395 \\
\bottomrule
\end{tabular}
\caption{Query formulation strategy comparison. Simple rewriting substantially outperforms more complex alternatives. Both domain-aware and multi-query strategies underperform even the no-rewriting baseline.}
\label{tab:strategies}
\end{table}

\subsection{Error Analysis}

Manual inspection of failure cases reveals three dominant error patterns.
First, \textbf{unanswerable queries} (${\sim}$25\% of the dataset) produce false positives when the system retrieves topically related but non-relevant passages, a limitation inherent to any retrieval-only system without answerability prediction.
Second, \textbf{long conversation histories} (10+ turns) may exceed the rewriter's context window (2048 tokens) in production settings; while the current dataset's conversations remain within this limit, longer real-world interactions would require truncation of early turns that may contain critical referents.
Third, FiQA is consistently the most challenging domain (nDCG@5 0.346 at best), attributable to its informal forum language, domain-specific financial jargon (\textit{``401k rollover''}, \textit{``FIRE movement''}), and abbreviations that create vocabulary mismatch for both lexical and semantic retrievers.

The substantial gap between development holdout performance (nDCG@5 0.422 at best uniform temperature) and official test set performance (0.531) suggests that domain-specific temperature tuning and the final pipeline configuration generalize well to unseen conversational patterns, and that our holdout set, which comprises only 22 conversations, may have been a conservative estimate of system capability.

% =============================================================================
% 6. CONCLUSION
% =============================================================================
\section{Conclusion}

We presented a three-stage retrieval pipeline for multi-turn conversational search that ranks 8th out of 38 systems on the MTRAGEval benchmark of SemEval 2026 Task 8 - Retrieval~\cite{Rosenthal2026MTRAGEval} with nDCG@5 of 0.531.
Each component (query rewriting, hybrid retrieval, and cross-encoder reranking) is justified through systematic comparison studies.

Our analysis surfaces two key insights: domain-specific temperature tuning for query generation yields meaningful gains by adapting to domain vocabulary characteristics, and larger reranking candidate pools counter-intuitively degrade quality.
Negative results with domain-aware prompting, multi-query expansion, and reranker fine-tuning reinforce that simplicity outperforms complexity when the base components are well-tuned.

Current limitations include English-only support, consumer hardware constraints that limited model sizes, and unweighted rank fusion.
Future work could explore larger rewriter models (14B+), learned fusion weights, ensemble reranking, cross-lingual extension, and a systematic study of the relationship between domain formality and optimal generation temperature for query rewriting.

% =============================================================================
% ACKNOWLEDGMENTS (does not count toward page limit)
% =============================================================================
\section*{Acknowledgments}
We thank the MTRAGEval organizers for creating a challenging and well-designed benchmark for multi-turn retrieval evaluation.

% =============================================================================
% REFERENCES (does not count toward page limit)
% =============================================================================
\bibliography{references}

% =============================================================================
% APPENDIX (does not count toward page limit)
% =============================================================================
\appendix

\section{Query Rewriting Prompt}
\label{app:prompt}

The following system prompt is used for query rewriting. The user message concatenates up to 10 turns of conversation history followed by the current question.

\begin{quote}
\small
\texttt{You are a query rewriting assistant for information retrieval. Given a conversation history and a current question, rewrite the question to be completely standalone and self-contained.}

\texttt{Rules:}\\
\texttt{1. Resolve all pronouns (it, they, this, that) to their explicit referents}\\
\texttt{2. Include relevant context from the conversation that's needed to understand the query}\\
\texttt{3. Keep the rewritten query concise and search-friendly}\\
\texttt{4. Do not add information not present in the conversation}\\
\texttt{5. If the question is already standalone, return it unchanged}
\end{quote}

%\section{Candidate Pool Size}
%\label{app:candidates}
%
%Table~\ref{tab:candidates} shows the effect of candidate pool size on reranking quality. Larger pools consistently degrade performance across all three models, as the cross-encoder assigns false-positive high scores to marginally relevant passages, displacing truly relevant documents from the final top-10.
%
%\begin{table}[h]
%\centering
%\small
%\begin{tabular}{lcccc}
%\toprule
%\textbf{Reranker} & $k{=}50$ & $k{=}100$ & $k{=}250$ & $k{=}500$ \\
%\midrule
%BGE-reranker-v2-m3 & \textbf{.416} & .400 & .390 & .388 \\
%ms-marco-MiniLM    & .375          & .373 & .369 & .366 \\
%BGE-base           & .307          & .290 & .266 & .251 \\
%\bottomrule
%\end{tabular}
%\caption{Effect of candidate pool size on reranking quality (nDCG@5). Larger pools consistently degrade performance across all three models.}
%\label{tab:candidates}
%\end{table}

\section{Hyperparameters}
\label{app:hyperparams}

Table~\ref{tab:hyperparams} lists all hyperparameters needed to reproduce our system.

\begin{table}[h]
\centering
\small
\begin{tabular}{ll}
\toprule
\textbf{Parameter} & \textbf{Value} \\
\midrule
\multicolumn{2}{l}{\textit{Query Rewriter (LoRA)}} \\
\quad Base model & Qwen2.5-7B-Instruct \\
\quad LoRA rank & 16 \\
\quad LoRA alpha ($\alpha$) & 32 \\
\quad LoRA dropout & 0.15 \\
\quad Target modules & q/k/v/o\_proj, \\
                     & gate/up/down\_proj \\
\quad Number of layers & 28 (all) \\
\quad Trainable params & 40.4M (0.53\%) \\
\quad Optimizer & AdamW \\
\quad Learning rate & $1 \times 10^{-5}$ \\
\quad Weight decay & 0.01 \\
\quad Batch size (micro) & 2 \\
\quad Gradient accumulation & 8 \\
\quad Effective batch size & 16 \\
\quad Training iterations & 500 \\
\quad Max sequence length & 2048 \\
\quad Gradient checkpointing & Yes \\
\quad Precision & bf16 \\
\quad Seed & 42 \\
\midrule
\multicolumn{2}{l}{\textit{BM25 Retrieval}} \\
\quad Library & BM25S \\
\quad Stopwords & English \\
\quad Stemming & None \\
\quad Candidates & 50 \\
\midrule
\multicolumn{2}{l}{\textit{Dense Retrieval}} \\
\quad Embedding model & BGE-base-en-v1.5 \\
\quad Embedding dim & 768 \\
\quad Index type & FAISS IndexFlatIP \\
\quad Normalization & L2 (cosine sim.) \\
\quad Candidates & 50 \\
\midrule
\multicolumn{2}{l}{\textit{Hybrid Fusion}} \\
\quad Method & RRF \\
\quad RRF $k$ & 60 \\
\midrule
\multicolumn{2}{l}{\textit{Cross-Encoder Reranking}} \\
\quad Model & BGE-reranker-v2-m3 \\
\quad Parameters & 568M \\
\quad Rerank candidates & 50 \\
\quad Output size & 10 \\
\quad Batch size & 8 \\
\midrule
\multicolumn{2}{l}{\textit{Domain Temperatures}} \\
\quad Cloud & 0.0 \\
\quad ClapNQ & 0.2 \\
\quad FiQA & 0.3 \\
\quad Govt & 0.3 \\
\bottomrule
\end{tabular}
\caption{Complete hyperparameter configuration for the final submission.}
\label{tab:hyperparams}
\end{table}

%\section{Detailed Per-Domain Results}
%\label{app:detailed}

% Table~\ref{tab:strategies} compares query formulation strategies.

%\begin{table}[h]
%\centering
%\small
%\begin{tabular}{lcc}
%\toprule
%\textbf{Strategy} & \textbf{nDCG@5} & \textbf{nDCG@10} \\
%\midrule
%No rewriting (baseline) & 0.371 & 0.412 \\
%\textbf{Simple rewriting ($t{=}0.2$)} & \textbf{0.422} & \textbf{0.467} \\
%Domain-aware prompting & 0.350 & 0.399 \\
%Multi-query expansion & 0.350 & 0.395 \\
%\bottomrule
%\end{tabular}
%\caption{Query formulation strategy comparison. Simple rewriting substantially outperforms more complex alternatives. Both domain-aware and multi-query strategies underperform even the no-rewriting baseline.}
%\label{tab:strategies}
%\end{table}

\end{document}